# Inference by Minimizing Size, Divergence, or their Sum

Sebastian Riedel    David A. Smith    Andrew McCallum
Department of Computer Science
University of Massachusetts
Amherst, MA 01003, U.S.A.
{riedel,dasmith,mccallum}@cs.umass.edu

## Abstract

We speed up marginal inference by ignoring factors that do not significantly contribute to overall accuracy. In order to pick a suitable subset of factors to ignore, we propose three schemes: minimizing the number of model factors under a bound on the KL divergence between pruned and full models; minimizing the KL divergence under a bound on factor count; and minimizing the weighted sum of KL divergence and factor count. All three problems are solved using an approximation of the KL divergence than can be calculated in terms of marginals computed on a simple seed graph. Applied to synthetic image denoising and to three different types of NLP parsing models, this technique performs marginal inference up to 11 times faster than loopy BP, with graph sizes reduced up to 98%—at comparable error in marginals and parsing accuracy. We also show that minimizing the weighted sum of divergence and size is substantially faster than minimizing either of the other objectives based on the approximation to divergence presented here.

## 1 INTRODUCTION

Computational resources are naturally scarce; hence, in practice, we often replace generic graphical models and their machinery with more restricted classes of distributions. For example, dependency parsing in Natural Language Processing can be formulated as inference in a Markov network (Smith & Eisner, 2008). While this approach allows great modelling flexibility and principled integration of several modules of an NLP pipeline, practical applications such as parsing web-scale data usually apply local classification approaches or models of limited structure that give better runtime guarantees than, say, BP on large and densely connected factor graphs (Nivre et al., 2004).

Here we show that generic approaches based on graphical models can be much more competitive if we restrict ourselves to a subgraph of the network that sufficiently approximates the full graph. Choi & Darwiche (2006) show how we can find such a graph by minimizing the KL divergence between the full and partial model; however, they require inference in the full graph—the very operation we want to avoid. How, then, can we efficiently find a sufficient subgraph *without* inference in the full network?

We show how—and when—the KL divergence can be approximated using the marginals of a simple proxy "seed" graph, such as a subgraph with only unary factors. We represent the divergence as the sum of three terms: (1) a sum over isolated per-factor divergences, based on the marginals of the seed graph, for all factors that are not in the target subgraph; (2) a term that measures the error we introduce by using the seed graph instead of the subgraph for calculating the divergence; and (3) a term that measures the error we introduce by considering the remaining factors in isolation. Informally, the error terms will disappear with increasing independence between the potentials of the remaining factors under the proxy distribution.

Based on this representation, we approximate the divergence by the sum of isolated per-factor gains. This approximation can be very efficiently computed but will work poorly if the unused potentials are highly correlated under the distribution of the proxy graph.

How can we use this approximation to decide which factors to ignore? We show three distinct ways to answer this question in a principled manner. First, we can minimize the size of the graph under a bound on the approximate KL divergence; second, we can minimize the approximate divergence under a bound on the number of factors; finally, we can minimize the weighted sum of both approximate divergence and factor count. The first and second objectives can be minimized in $O(n \log n)$ time, where $n$ is the number of factors that are not in the proxy graph. The third objective only requires $O(n)$ time.

We empirically evaluate all approaches on state-of-the-art graphical models for dependency parsing in several languages. For some of these models no efficient dynamic programs (DP) exist. For others, DPs are available, but belief propagation (BP) has been shown to be significantly faster while providing comparable accuracy (Smith & Eisner, 2008).

We show that our three objectives speed up BP inference by an order of magnitude with no loss in application accuracy (e.g., 11 times for English). This is achieved while using only a fraction (e.g., 1.3% for English) of the factors in the original graph. We also show that jointly minimizing the sum of size and divergence is the fastest option, with accuracy comparable to the other two objectives.

In the following, we first introduce our notation (§2). Then we present the optimization problems we are trying to solve and upper bounds on the KL divergence that make this possible (§3). We relate our approach to existing work (§4), evaluate it on a synthetic image denoising and real-world NLP data (§5), and conclude (§6). In the appendix we give proof sketches of our approximations and bounds.

## 2 BACKGROUND

Here we consider binary Markov networks (Wainwright & Jordan, 2008): our distributions are represented by undirected graphical models over the space $\mathcal{Y} \stackrel{\text{def}}{=} \{0,1\}^{|V|}$ of assignments to binary variables $V$:

$$p_{\mathcal{F}}(\mathbf{y}) \stackrel{\text{def}}{=} \frac{1}{Z} \prod_{i \in \mathcal{F}} \Psi_i(\mathbf{y}). \quad (1)$$

specified by an index set $\mathcal{F}$ and a corresponding family $(\Psi_i)_{\mathcal{F}}$ of factors $\Psi_i : \mathcal{Y} \mapsto \Re^+$. Here $Z$ is the partition function $Z_{\mathcal{F}} = \sum_{\mathbf{y} \in \mathcal{Y}} \prod_i \Psi_i(\mathbf{y})$. We will restrict our attention to binary factors that can be represented as $\Psi_i(\mathbf{y}) \stackrel{\text{def}}{=} e^{\theta_i \phi_i(\mathbf{y})}$ with binary feature functions $\phi_i(\mathbf{y}) \in \{0,1\}$. Finally, for a subset of factors $\mathcal{G} \subseteq \mathcal{F}$ we will write $\Psi_{\mathcal{G}}$ to mean $\prod_{i \in \mathcal{G}} \Psi_i$.

When using Markov networks we often need to calculate expectations of the variables and features under the model. These can be used for calculating the gradient during maximum likelihood learning and providing confidence values for predictions. They are also often used for posterior decoding, where the state of each variable is chosen to maximize its marginal probability, possibly subject to constraints.

Formally, we seek to find the mean vector of the sufficient statistics of $p_{\mathcal{F}}$. That is, for each factor $i$ we want to calculate

$$\mu_i^{\mathcal{F}} \stackrel{\text{def}}{=} \sum_{\phi_i(\mathbf{y})=1} p_{\mathcal{F}}(\mathbf{y}) = \mathrm{E}_{\mathcal{F}}[\phi_i]. \quad (2)$$

Note that the expectations of a variable can be formulated as the expectation of a unary feature for this variable. We also often need the expectations and covariances for potentials and their products, abbreviated as $\mathrm{E}_X^Y \stackrel{\text{def}}{=} E_X[\Psi_Y]$ and $\mathrm{Cov}_Z^{X,Y} \stackrel{\text{def}}{=} \mathrm{Cov}_Z(\Psi_X, \Psi_Y)$

We assume that we already have some means of inference for our model at hand—at least for small versions of it. This could be, among others, loopy belief propagation (Murphy et al., 1999), a Gibbs sampler (Geman & Geman, 1990), a naive or structured mean field approach (Jordan et al., 1999), or a further developed and optimized version of the above.

Many of these approximations, however, are still too slow for practical purposes because they scale, at best, linearly with network size. When we consider factor sets that grow super-exponentially (Culotta et al., 2007), or at least with a high polynomial (e.g., when grounding factor templates with many free variables as present in Markov Logic Networks), inference soon becomes the bottleneck of our application. But do these networks need to be so large, or can we ignore factors, speed-up inference, and still achieve sufficient accuracy?

## 3 MINIMIZING SIZE AND DIVERGENCE

As noted above, there are least three ways to tackle the question of finding an optimal subset of factors to approximate the full network. The first can be phrased as follows: among the distributions that are within some bounded KL divergence to the full graph, find the one that is smallest in terms of factor count. Formally, we need to find $p_{\hat{\mathcal{H}}}$ such that

$$\hat{\mathcal{H}} = \underset{\mathcal{H} \subset \mathcal{F}, D(p_{\mathcal{H}}||p_{\mathcal{F}}) \leq \epsilon}{\arg \min} |\mathcal{H}|. \quad (3)$$

The second way amounts to the following question: among the family of distributions that arise from picking $m$ factors out of the full graph, which one is the closest to the full distribution? Formally, we have to find $p_{\hat{\mathcal{H}}}$ such that

$$\hat{\mathcal{H}} = \underset{=\mathcal{H} \subset \mathcal{F}, |\mathcal{H}| \leq m}{\arg \min} D(p_{\mathcal{H}}||p_{\mathcal{F}}). \quad (4)$$

This is formulation is variational in the traditional sense: we are searching for a good approximation in a

family of tractable distributions, where "tractable" is defined as "small in size."

Finally, we can ask: what is the sub-graph that minimizes the weighted sum of size and KL divergence?

$$\hat{\mathcal{H}} = \arg\min_{\mathcal{H} \subset \mathcal{F}} D\left(p_{\mathcal{H}}||p_{\mathcal{F}}\right) + \gamma \left|\mathcal{H}\right|. \qquad (5)$$

In this case we are explicitly searching for a trade-off between efficiency and accuracy and do not need to commit to a specific bound on size or divergence. Crucially, it turns out that for the approximations on the KL divergence presented here, choosing $\hat{\mathcal{H}}$ for problem 5 is asymptotically and empirically faster than for problems 3 and 4.

All three problems are combinatorial—the space of models we have to search is exponential in $|\mathcal{F}|$. Obviously, we cannot search it exhaustively, not only because the space is large, but also because evaluating the KL divergence generally requires inference in $p_{\mathcal{H}}$.

### 3.1 KL DIVERGENCE WITH PROXIES

We cannot try every possible subset of factors $\mathcal{H}$, perform inference with it, and calculate its KL divergence to $\mathcal{F}$. Instead, we can use some small initial seed graph, perform inference in it, and use the results to approximate the KL divergences for several $\mathcal{H}$. Clearly, the effectiveness of this approach hinges on the choice of seed graph, but previous work in relaxation and cutting plane methods has shown that simple "local" approximations can go a long way as initial formulations of inference problems (Riedel, 2008; Tromble & Eisner, 2006; Anguelov et al., 2004). In the following we will qualify and quantify when this approximation works well.

The cornerstone of our approach is a formulation of the KL divergence for $p_{\mathcal{H}}$ that factors into isolated per-factor divergences which can be calculated using the beliefs of a *proxy* graph $\mathcal{G} \subset \mathcal{H}$. In the following we will assume that $\mathcal{N} \stackrel{\text{def}}{=} \mathcal{H} \setminus \mathcal{G}$ and $\mathcal{R} \stackrel{\text{def}}{=} \mathcal{F} \setminus \mathcal{H}$; that is, $\mathcal{H}$ can be derived from the proxy $\mathcal{G}$ by adding the *new* factors $\mathcal{N}$, and from the full graph $\mathcal{H}$ by removing the *remaining* factors $\mathcal{R}$.

**Proposition 1.** *Let* $S_Z^{X,Y} \stackrel{\text{def}}{=} \log\left(\frac{\text{Cov}_Z^{X,Y}}{E_Z^X E_Z^Y} + 1\right) - \frac{\text{Cov}_Z(\log(\Psi_X), \Psi_Y)}{E_Z(\log(\Psi_X))}$ *and* $I_Z^X \stackrel{\text{def}}{=} \log\left(\frac{E_Z^X}{\prod_{i \in X} E_Z^i}\right)$, *then*

$$D\left(p_{\mathcal{H}}||p_{\mathcal{F}}\right) = \sum_{i \in \mathcal{R}} D\left(p_{\mathcal{G}}||p_{\mathcal{G} \cup \{i\}}\right) + S_{\mathcal{G}}^{\mathcal{N},\mathcal{R}} + I_{\mathcal{G}}^{\mathcal{R}}.$$

In words: we can calculate the divergence between $p_{\mathcal{H}}$ and $p_{\mathcal{F}}$ by first summing over the divergences between $p_{\mathcal{G}}$ and $p_{\mathcal{G} \cup \{i\}}$ for all remaining factors $i \in \mathcal{R}$, and then adding two error terms: $S_{\mathcal{G}}^{\mathcal{N},\mathcal{R}}$ that measures a correlation between the new product of potentials $\Psi_{\mathcal{N}}$ and remaining product $\Psi_{\mathcal{R}}$; and $I_{\mathcal{G}}^{\mathcal{R}}$ that measures a correlation between the remaining individual potentials $\Psi_i$ with $i \in \mathcal{R}$.

This leads us to the following *factorized proxy-based* approximation of the KL divergence

$$D_{\mathcal{G}}^1\left(p_{\mathcal{H}}||p_{\mathcal{F}}\right) \stackrel{\text{def}}{=} \sum_{i \in \mathcal{R}} D\left(p_{\mathcal{G}}||p_{\mathcal{G} \cup \{i\}}\right). \qquad (6)$$

It is this approximation that we will consider when minimizing the three objectives we have presented earlier. Clearly, its accuracy will depend on the correlation between the features of the factors in $\mathcal{R}$ and $\mathcal{N}$, as measured by $S_{\mathcal{N},\mathcal{R}}^{\mathcal{G}} + I_{\mathcal{R}}^{\mathcal{G}}$. In particular, it is easy to show that if the features of all remaining factors $\mathcal{R}$ are independent of each other, and if $\Psi_{\mathcal{N}}$ is independent of $\Psi_{\mathcal{R}}$, then the approximation is exact.

Finally, note that Riedel & Smith (2010) show that with an expectation for the feature $\phi_i$ under $p_{\mathcal{G}}$ we can efficiently calculate the *gain* $D\left(\mathcal{G}||\mathcal{G} \cup \{i\}\right)$ as

$$g_{\mathcal{G}}(i) \stackrel{\text{def}}{=} D\left(p_{\mathcal{G}}||p_{\mathcal{G} \cup \{i\}}\right) = \log\left(1 - \mu_i^{\mathcal{G}} + \mu_i^{\mathcal{G}} e^{\theta_i}\right) - \mu_i^{\mathcal{G}} \theta_i.$$

### 3.2 SIZE MINIMIZATION

The factorized proxy-based divergence allows us to find an approximation to problem 3 by simply minimizing the number $|\mathcal{N}|$ of new factors under a constraint on the approximate divergence $D_{\mathcal{G}}^1$. That is, for a given proxy $\mathcal{G}$ we need to find the additional factors $\mathcal{N}$ according to

$$\arg\min_{\mathcal{N}: D_{\mathcal{G}}^1(p_{\mathcal{G} \cup \mathcal{N}}||p_{\mathcal{F}}) \leq \epsilon} |\mathcal{N}|. \qquad (7)$$

This can be achieved by first sorting the factors according to their gain $g_{\mathcal{G}}(i)$. Then we iterate over the sorted factors, starting at the lowest gain, and discard each factor $i$ until the sum of the gains of all discarded factors exceeds $\epsilon$.

### 3.3 DIVERGENCE MINIMIZATION

The factorized proxy-based approximation also helps us to find an approximate solution to problem 8 by solving

$$\arg\min_{\mathcal{N}: |\mathcal{N}| \leq m - |\mathcal{G}|} D_{\mathcal{G}}^1\left(p_{\mathcal{G} \cup \mathcal{N}}||p_{\mathcal{F}}\right). \qquad (8)$$

This amounts to simply choosing the $m - |\mathcal{G}|$ factors $i$ with highest $g_{\mathcal{G}}(i)$, which again requires sorting the candidate factors.

**Algorithm 1** Ignorant Inference using black-box solver $S$ and initial seed graph $\mathcal{G}$.

1:    $\mu \leftarrow \text{infer}_S(\mathcal{G})$
       *pick additional factors w.r.t. to objective and bound*
2:    $\mathcal{N} \leftarrow \text{pickUntilSumExceeds}_\epsilon(\mu, \mathcal{F} \setminus \mathcal{G})$
      $\mathcal{N} \leftarrow \text{pickUntilCountExceeds}_m(\mu, \mathcal{F} \setminus \mathcal{G})$
      $\mathcal{N} \leftarrow \text{pickAllOver}_\gamma(\mu, \mathcal{F} \setminus \mathcal{G})$
       *add to seed graph*
3:    $\mathcal{H} \leftarrow \mathcal{G} \cup \mathcal{N}$
       *inference on final model, if changed*
4:    **if** $\mathcal{H} \neq \mathcal{G}$   $\mu \leftarrow \text{infer}_S(\mathcal{H})$

### 3.4 JOINT MINIMIZATION

Based on the approximate divergence we can approximate the objective in problem 5 by $\gamma|\mathcal{F}| + \sum_{i \in \mathcal{F} \setminus \{\mathcal{G} \cup \mathcal{N}\}} (g_\mathcal{G}(i) - \gamma)$. This leads to the equivalent optimization problem

$$\arg\min_\mathcal{N} \sum_{i \in \mathcal{F} \setminus \{\mathcal{G} \cup \mathcal{N}\}} (g_\mathcal{G}(i) - \gamma) \qquad (9)$$

This problem has the additional benefit of not requiring sorting of candidate factors: in order to solve it we simply add all factors with $g_\mathcal{G}(i) \geq \gamma$. Choosing $\mathcal{H}$ thus requires $O(|\mathcal{F} \setminus \mathcal{G}|)$ rather than $O(|\mathcal{F} \setminus \mathcal{G}| \log |\mathcal{F} \setminus \mathcal{G}|)$ time.

### 3.5 ALGORITHM

Algorithm 1 summarizes our method and the three basic variations we investigate here. We start with inference in some initial graph $\mathcal{G}$ (step 1) using some black-box method $S$. In our case this graph is always the fully factorized set of local factors. Then we choose an additional set of factors $\mathcal{N}$ either by discarding them until the sum of discarded factor gains $g_\mathcal{G}(i)$ exceeds $\epsilon$ (MinSize), by picking the $m$ factors with highest gain (MinDivergence), or by picking all factors with gain higher than $\gamma$ (MinJoint).

### 3.6 UPPER BOUND FOR DIVERGENCE

Once we have found an $\mathcal{H}$ and performed inference in it, we can ask how close it really is to the full graph $\mathcal{F}$ in terms of KL divergence. This could be used to decide whether to continue inference, or to give confidence intervals for the marginals we return. Note that here we do not require a proxy graph; instead, we can directly evaluate $D(\mathcal{H}||\mathcal{F})$ using the inference result of the algorithm. Also note that the bounds in §3.1 cannot be applied here because we cannot easily evaluate the constants that depend on second order moments.

In the following we give the tightest possible bound when we only want to consider first order moments (i.e., feature expectations). This bound is derived from an exact bound on the expectation presented by Kreinovich et al. (2003) and assumes, with no loss of generality, positive weights.

**Proposition 2.** *Let $L \overset{def}{=} |\mathcal{F} \setminus \mathcal{H}|$ and $(i_1, \ldots, i_L)$ be a sequence of the remaining factors $\mathcal{F} \setminus \mathcal{H}$ sorted in increasing order of means $\mu_i^\mathcal{H}$. Let $\mu'_0 \overset{def}{=} 0$, $\mu'_{L+1} \overset{def}{=} 1$ and $\mu'_j \overset{def}{=} \mu_{i_j}^\mathcal{H}$ for $j \in \{1, \ldots, L\}$. With $C \overset{def}{=} \log\left(\sum_{j=0}^L (\mu'_{j+1} - \mu'_j) e^{-\sum_{k=1}^j \theta_{i_k}}\right)$ we then have*

$$D(p_\mathcal{H}||p_\mathcal{F}) \leq \|\theta_{\mathcal{F} \setminus \mathcal{H}}\|_1 - \langle \mu_{\mathcal{F} \setminus \mathcal{H}}^\mathcal{G}, \theta_{\mathcal{F} \setminus \mathcal{H}} \rangle - C$$

*and there exists at least one distribution with the given means for which the divergence equals the upper bound.*

Riedel & Smith (2010) give a looser version of this bound without the C term (which is always non-negative). Also, while this bound is the tightest possible when we are only given potential expectations, it may still be quite loose in many cases. This is due to the fact that it allows potentials to be maximally correlated and in practice this is rarely the case.

## 4 RELATED WORK

There is a link between our work and relaxation approaches for MPE inference, in which violated factors/constraints of a graphical model (Anguelov et al., 2004; Riedel, 2008), weighted Finite State Machine (Tromble & Eisner, 2006) or Integer Linear Program (Riedel & Clarke, 2006) are incrementally added to a partial graph. Our work differs in that we (a) tackle marginal instead of MPE inference and (b) frame the problem of choosing factors as optimization problem.

Sontag & Jaakkola (2007) compute marginal probabilities by using a cutting plane approach that starts with the local polytope and then optimizes some approximation of the log partition function. Hence they ignore constraints instead of factors. While this approach does tackle marginalization, it is focusing on improving its accuracy. In particular, the optimization problems they solve in each iteration are in fact larger than the problem we want to compress (if we use BP as solver for our sub-problems).

This paper can partly be seen as theoretical justification for the work of Riedel & Smith (2010), who heuristically use a threshold on the gain function (§3.1) in each iteration in order to choose which factors to add. Here we prove that they are effectively optimizing the weighted sum of proxy-based factorized divergence and graph size we present in §3.4. Moreover, we present a tighter bound on divergence—in fact, the tightest bound possible when we only first order moments are available. We also provide further experiments and analysis. Previous work on training-time feature induction for MRFs (as well as decision trees) also applied greedy, iterative divergence minimization (Della Pietra et al., 1997), and it would be interesting to apply our methods for adding multiple features at training time.

Our approach is also closely related to traditional variational methods such as the naive and structured mean field (Wainwright & Jordan, 2008). While these approaches assume the structure to be fixed and search for good parametrization/weights, we keep the parametrization fixed and search for good structure. One advantage of our approach is that it can dynamically adapt to the complexity of the model. That is, while some some areas of the graph may be modelled as well through local factors alone, some others are easiest to capture by the (possibly loopy) structure of the full graph. Where these areas are usually depends on the observed evidence, and hence we cannot identify a fixed structure in advance. Moreover, in our case we do not pick a particular family of distributions in advance; instead, we allow it to grow as needed.

Finally, our approach is related to edge deletion in Bayesian networks (Choi & Darwiche, 2006). They remove edges from a Bayesian network in order to find a close approximation to the full network useful for other inference-related tasks (such as MAP inference). The most obvious difference is the use of Bayesian Networks instead of Markov Networks. They also ask which factors to *remove* from the full graph, instead of which to *add* to a partial graph. This requires inference in the full model—the very operation we want to avoid. Moreover, the method they use for choosing edges to remove corresponds to MinDivergence, and therefore requires sorting of edges. Hence our MinJoint approach is asymptotically faster.

## 5 RESULTS

We now investigate and describe our algorithms in a controlled environment to display some of their properties (§5.1) and show their effectiveness in a real-world NLP application (§5.2).

### 5.1 SYNTHETIC DATA

Our approach is inspired by the following observation we made in practice: Often models consists of a very confident "local" graph, and the long distance factors only serve to correct the few mistakes the local model makes. We argue that in such cases (a) the three objectives choose a small subset of factors that is sufficient to correct the remaining errors, and (b) the decrease in graph size leads to quicker inference.

To test this claim in a controlled environment we use an existing synthetic image denoising dataset (Kumar & Hebert, 2004). It consists of 4 original images, and 50 versions of each with added noise. The factor graph we apply is an Ising model on a 64 x 64 binary grid with nodes $V$ and edges $E$ defined as follows:

$$p(\mathbf{y}|\mathbf{x}) = \frac{1}{Z_\mathbf{x}} \exp\left(\sum_{(i,j)\in E} [y_i = y_j] + \alpha \sum_{i\in V} x_i y_i\right)$$

Here each $x_i$ is the observed pixel value with added noise, normalized to have 0 mean. The parameter $\alpha$ controls the confidence we give into the unary factors, and the observations we have. Hence, if we choose the unary factors as seed graph, we hope that for large $\alpha$ our algorithms will add only a few factors but still produce accurate results. Note that this model uses the edge weight 1. We ran experiments with varying weights and observed the same qualitative trends.[1]

We choose the free parameters $m$, $\epsilon$ and $\gamma$ for the different optimization strategies by setting $\alpha = 5$ as reference point, and tune each parameter to yield factor graphs with 50% of the original factors. This allows us to directly compare the different speed-ups and errors at $\alpha = 5$, as well as each methods qualitative behaviour for a range of $\alpha$ values. Finally, for inference in the sub-graphs, and full graph, we use Tree-Reweighted BP (TRW).

Figure 1a shows the relative size of the final factor graph compared to the full graph for our three objectives. Figure 1b presents the corresponding average error in marginals with respect to the marginals of the full graph (approximately calculated using TRW). Finally, figure 1c shows the relative speed-ups compared to inference in the full graph.

All figures show a clear picture: the more confidence we have in our seed graph, the better all methods perform. For the case of MinDivergence (where we bound size) we achieve increasingly accurate results with the same number of factors. For both MinSize (where we bound divergence) and MinJoint we require fewer factors to achieve equivalent levels of accuracies. Moreover, this reduction in size leads to direct improvements in inference speed, despite having to run inference twice.

Finally, note that for $\alpha = 5$ (all methods give 50% reductions in graph size) MinJoint is clearly faster than both other approaches. This suggests that its advan-tage in terms of asymptotic complexity pays off in this case. We make a similar observation for parsing below.

---

[1] Clearly inference in general became harder with higher interaction strength. We omit further details for brevity.

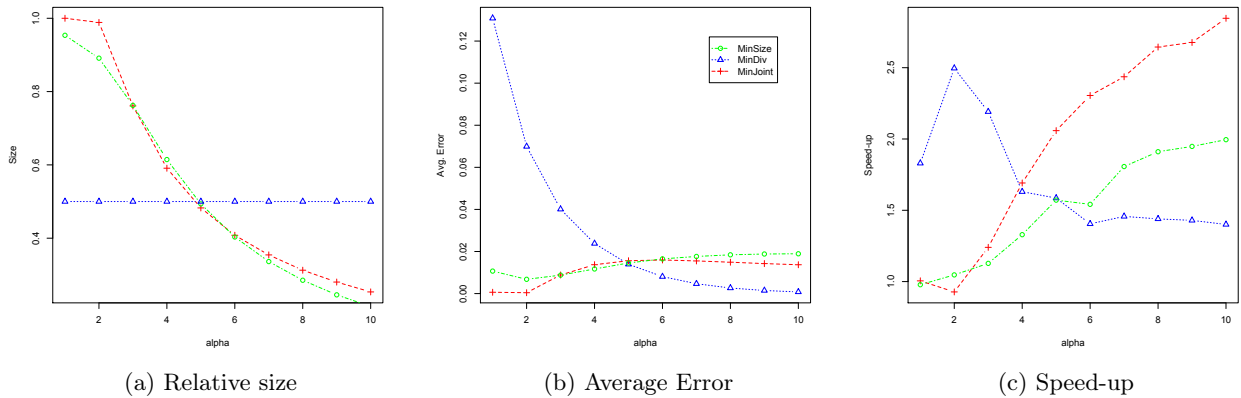

(a) Relative size     (b) Average Error     (c) Speed-up

Figure 1: Impact of varying confidence parameter $\alpha$.

## 5.2 DEPENDENCY PARSING

The synthetic nature of the above experiments helped us to describe some of the basic properties of MinSize, MinJoint and MinDivergence, but it is not yet clear whether these are useful methods in practice. To show their utility in the real world, we also evaluate them on three second order dependency parsing models, as presented by Smith & Eisner (2008). These models yield state-of-the-art performance for many languages and are very relevant in practice.

These factor graphs are computationally challenging: they employ a number of variables quadratic in the sentence length to represent each of the candidate dependency links between words, and a cubic number of pairwise factors between them. Factor graphs for English, for example, average about 700 nodes and 20,000 factors per sentence.

The models we consider here also contain a single "combinatorial factor" that connects all $n^2$ variables and ensures that the distribution models spanning trees over the tokens of the sentence. Since there are an exponential number of valid trees, naively computing this factor's messages is intractable; instead, we follow Smith & Eisner (2008) in calculated messages by combinatorial optimization. Crucially, our approach treats this machinery to calculate the initial and final beliefs (steps 2 and 4 of algorithm 1) as a black box.

We trained and tested on a subset of languages from the CoNLL Dependency Parsing Shared Tasks (Nivre et al., 2007): Dutch, Italian, and English. Our models differ in the number and type of edges (only grandparent edges for Dutch and Italian, grandparent and sibling edges for English), and in the type of combinatorial factor used to ensure the proper tree structure for the language at hand (non-projective for Dutch, projective for Italian and English). Our models would rank highly among the shared task submissions, but could surely be further improved.

We evaluate the accuracy of inference in two ways. First, we measure the error of our marginals relative to the marginals calculated by full BP. For each sentence, we find the node with highest error and then average these maximal errors over the corpus.

Absolute errors on marginals do not give the full picture. One way to measure the effective accuracy in practice is to use posterior decoding and find the tree that maximizes the calculated marginal beliefs. This allows us to compare against a gold standard tree.

### 5.2.1 Averaged Performance

The results in terms of average dependency accuracy, speed, and final factor graph size can be seen in table 1. We compare full BP (with maximally 10 iterations) to the three setups we present in §3. For each method and language we find the most efficient parameters $\epsilon$, $m$ and $\gamma$ that lead to the same dependency accuracy as BP in the full graph, if possible in a reasonable amount of time. See §5.2.3 for a closer look.

The table shows that, across the board, a significant reduction in size is possible without loss of dependency accuracy. Moreover, for the case of Italian we can calculate the marginals exactly and hence measure the true error of both BP and our approximations. Note that they are in fact nearly identical.

Clearly, the reduced size has a direct effect on speed. For example, for the MinJoint algorithm we observe a 4 times (Dutch), 7 times (Italian), and 11 times (English) speed-up. MinJoint outperforms both other approaches, with comparable accuracy at higher speed.

Table 1: Averaged results at same dependency accuracy; F=Percentage of binary edges added; T=time in ms; E=averaged max. error in marginals; A=accuracy on gold standard; NP=Non-projective; P=projective; GP=grandparent edges; SIB=sibling edges. *E=calculated w.r.t. to the true marginals as calculated by DP.

| Conf. | Dutch(NP,GP) | | | | Italian(P,GP) | | | | English (P, GP+SIB) | | | |
|---|---|---|---|---|---|---|---|---|---|---|---|---|
| | F | T | E | A | F | T | E* | A | F | T | E | A |
| BP | 100 | 350 | 0 | 84.0 | 100 | 890 | 14.5 | 86.9 | 100 | 2370 | 0 | 88.4 |
| MinDivergence | 21.5 | 121 | 1.4 | 83.9 | 3.4 | 162 | 14.6 | 86.9 | 1.5 | 248 | 14.1 | 88.5 |
| MinSize | 23.0 | 141 | 1.8 | 83.9 | 4.0 | 161 | 14.7 | 87.0 | 1.4 | 245 | 16.5 | 88.5 |
| MinJoint | 12.3 | 79 | 1.8 | 84.0 | 1.8 | 129 | 14.7 | 87.0 | 1.3 | 210 | 16.1 | 88.5 |

#### 5.2.2 Scaling

Figure 2a shows the number of added factors versus the number of nodes in a graph for Italian. (Other languages showed the same trends.) As expected, the graph size stays constant for MinDivergence inference, which means that we will allocate too many factors for small problems if we want to allocate enough for large ones. By contrast, the factor count required by MinSize and MinJoint adapt with problem size.

Figure 2b shows that runtime also scales better with graph size. In fact, for MinJoint, MinSize and MinDiverence we observe a linear behaviour with the number of nodes, akin to the observations of Riedel (2008) in relaxed/cutting plane MAP inference. Asymptotically, however, our algorithm still scales with the number of factors $|\mathcal{F}|$ (or $|\mathcal{F}|\log|\mathcal{F}|$ when sorting) since we have to scan all remaining factors as candidates.

#### 5.2.3 Choosing Factors

Which scheme should be chosen to pick factors? The previous sections suggest that MinJoint is the better approach because it is faster. To show that this observation is not based on better tuning of the parameter $\gamma$, we show how each method performs when we vary their free parameters. Figure 2c presents the effectiveness of each scheme: For several values of the parameter $\epsilon$ (when minimizing size), $m$ (when minimizing divergence) or $\gamma$ (when minimizing both) we draw average time spent vs average accuracy.

We notice that MinJoint makes significantly better use of its runtime. In particular, we see that MinSize and MinDivergence cannot be configured to achieve the accuracy of BP (at 0.15) as quickly as MinJoint.

### 6 CONCLUSION

We have shown how to substantially speed up marginal inference by ignoring a set of factors that do not significantly contribute to overall accuracy. We surely cannot expect this always to work well: we require factor graphs that contain small but highly confident sub-models that can be used as initial proxy graphs (compare §5.1). However, we think that in many cases local clues are enough to determine the state of a variable and can hence be incorporated into powerful local sub-graphs. This has been frequently observed in NLP, where local features are in fact often hard to beat (Bengtson & Roth, 2008). It is also crucial for the success of factor reduction methods in MPE inference (Riedel, 2008; Tromble & Eisner, 2006) for tasks such as entity resolution or semantic role labeling.

We want to extend our work in two directions. First, we want to ignore not only factors, but also values. We believe that in many cases models have an almost deterministic belief about the state of certain variables, and in these cases we can gain significant efficiencies by ignoring the other states entirely. Second, we want to investigate reparametrization of the remaining factors in the spirit of Choi & Darwiche (2006).

### Appendix: Proof Sketches

Proposition 1 relies on the fact that we can represent $p_{\mathcal{G}\cup\mathcal{N}}(\mathbf{y})$ using $Z_{\mathcal{G}\cup\mathcal{N}}^{-1} Z_{\mathcal{G}} \Psi_{\mathcal{N}}(\mathbf{y}) p_{\mathcal{G}}(\mathbf{y})$ and calculate the quotient of partition functions using

$$Z_{\mathcal{G}\cup\mathcal{N}}^{-1} Z_{\mathcal{G}} = \mathrm{E}_{\mathcal{G}}\left[\Psi_{\mathcal{N}}\right]^{-1}. \qquad (10)$$

Hence we can evaluate expectations under $\mathcal{H} = \mathcal{G}\cup\mathcal{N}$ based on expectations under $\mathcal{G}$ using

$$\mathrm{E}_{\mathcal{G}\cup\mathcal{N}}\left[x(\mathbf{y})\right] = \mathrm{E}_{\mathcal{G}}\left[x(\mathbf{y})\Psi_{\mathcal{N}}\right]\mathrm{E}_{\mathcal{G}}\left[\Psi_{\mathcal{N}}\right]^{-1}. \qquad (11)$$

Using the primal form of the KL divergence Wainwright & Jordan (2008) (slightly adapted) we get

$$D\left(p_{\mathcal{H}}||p_{\mathcal{F}}\right) = \log\left(Z_{\mathcal{F}}Z_{\mathcal{G}\cup\mathcal{N}}^{-1}\right) - \mathrm{E}_{\mathcal{H}}\left(\log\left(\Psi_{\mathcal{F}\setminus\mathcal{H}}\right)\right).$$

By plugging in 10 and 11 we can write

$$D\left(p_{\mathcal{H}}||p_{\mathcal{F}}\right) = \log\left(\frac{\mathrm{E}_{\mathcal{G}}^{\mathcal{R}\cup\mathcal{N}}}{\mathrm{E}_{\mathcal{G}}^{\mathcal{N}}}\right) - \frac{\mathrm{E}_{\mathcal{G}}\left[\log\left(\Psi_{\mathcal{R}}\right)\Psi_{\mathcal{N}}\right]}{\mathrm{E}_{\mathcal{G}}^{\mathcal{N}}}.$$

Setting $\alpha \stackrel{\text{def}}{=} \frac{\mathrm{E}_{\mathcal{G}}^{\mathcal{R}\cup\mathcal{N}}}{\mathrm{E}_{\mathcal{G}}^{\mathcal{R}}\mathrm{E}_{\mathcal{G}}^{\mathcal{N}}}$ and $\beta \stackrel{\text{def}}{=} \mathrm{E}_{\mathcal{G}}\left[\log\left(\Psi_{\mathcal{R}}\right)\Psi_{\mathcal{N}}\right] - \mathrm{E}_{\mathcal{G}}\left[\log\left(\Psi_{\mathcal{R}}\right)\right]\mathrm{E}_{\mathcal{G}}^{\mathcal{N}}$, we get $\log\left(\mathrm{E}_{\mathcal{G}}^{\mathcal{R}}\right) - \mathrm{E}_{\mathcal{G}}\left[\log\left(\Psi_{\mathcal{R}}\right)\right] + \log\left(\alpha\right) - \frac{\beta}{\mathrm{E}_{\mathcal{G}}^{\mathcal{N}}}$.

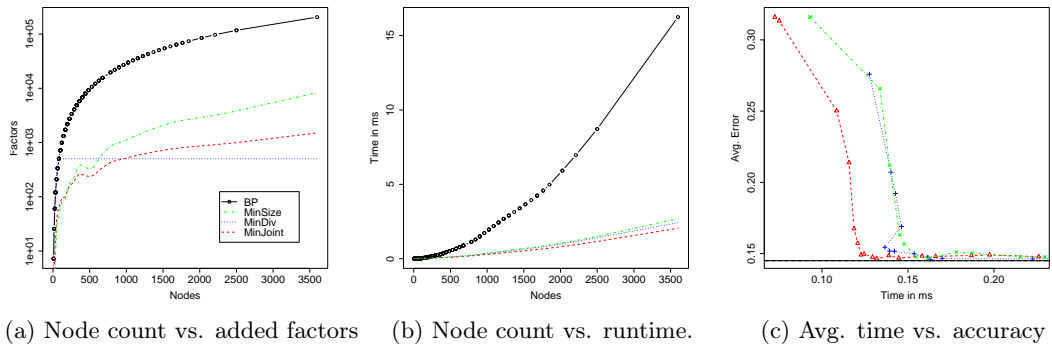

(a) Node count vs. added factors    (b) Node count vs. runtime.    (c) Avg. time vs. accuracy

Figure 2: Runtime, factor count and dependency accuracy.

With $\eta \stackrel{\text{def}}{=} \frac{E_{\mathcal{G}}^{\mathcal{R}}}{\prod_{i \in \mathcal{R}} E_{\mathcal{G}}^{i}}$ we can replace the first term by $\sum_{i \in \mathcal{R}} \log\left(E_{\mathcal{G}}^{i}\right) + \log\left(\eta\right)$. The second term can we be rewritten as $-\sum_{i \in \mathcal{R}} \mu_i^{\mathcal{G}} \theta_i$ using the additivity of expectations.

We can hence write the divergence as $\sum_{i \in \mathcal{R}} \log\left(E_{\mathcal{G}}^{i}\right) - \mu_i^{\mathcal{G}} \theta_i + \log\left(\alpha\eta\right) - \frac{\beta}{E_{\mathcal{G}}^{\mathcal{N}}}$. Expanding $\alpha$, $\beta$ and $\eta$ while using the definition of covariance and replacing $\log\left(E_{\mathcal{G}}^{i}\right) - \mu_i^{\mathcal{G}} \theta_i$ by $D\left(p_{\mathcal{G}} || p_{\mathcal{G} \cup \{i\}}\right)$ leads to the desired representation.

For proposition 2 we use

$$D\left(p_{\mathcal{H}} || p_{\mathcal{F}}\right) = \log\left(E_{\mathcal{G}}^{\mathcal{F} \setminus \mathcal{H}}\right) - E_{\mathcal{G}}\left[\log\left(\Psi_{\mathcal{F} \setminus \mathcal{H}}\right)\right]$$

and apply the bound of Kreinovich et al. (2003) for expectations of products of variables to $E_{\mathcal{G}}^{\mathcal{F} \setminus \mathcal{H}}$.

## Acknowledgements

This work was supported in part by the Center for Intelligent Information Retrieval, in part by SRI International subcontract #27-001338 and ARFL prime contract #FA8750-09-C-0181, in part by Army prime contract number W911NF-07-1-0216 and University of Pennsylvania subaward number 103-548106, and in part by UPenn NSF medium IIS-0803847. Any opinions, findings and conclusions or recommendations expressed in this material are the authors' and do not necessarily reflect those of the sponsor.

## References

Anguelov, D., Koller, D., Srinivasan, P., Thrun, S., Pang, H.-C., and Davis, J. The correlated correspondence algorithm for unsupervised registration of nonrigid surfaces. In *NIPS*, 2004.

Bengtson, E. and Roth, D. Understanding the value of features for coreference resolution. In *EMNLP*, 2008.

Choi, A. and Darwiche, A. A variational approach for approximating bayesian networks by edge deletion. In *UAI*, 2006.

Culotta, A., Wick, M., Hall, R., and McCallum, A. First-order probabilistic models for coreference resolution. In *HLT-NAACL*, 2007.

Della Pietra, S., Della Pietra, V., and Lafferty, J. Inducing features of random fields. *IEEE Trans. Pattern Analysis and Machine Intelligence*, 19(4):380–393, 1997.

Geman, S. and Geman, D. Stochastic relaxation, Gibbs distributions, and the Bayesian restoration of images. In *Readings in Uncertain Reasoning*, pp. 452–472. Morgan Kaufmann, 1990.

Jordan, M. I., Ghahramani, Z., Jaakkola, T., and Saul, L. K. An introduction to variational methods for graphical models. *Machine Learning*, 37(2):183–233, 1999.

Kreinovich, V., Ferson, S., and Ginzburg, L. Exact upper bound on the mean of the product of many random variables with known expectations. *Reliable Computing*, 9(6): 441–463, 2003.

Kumar, S. and Hebert, M. Approximate parameter learning in discriminative fields. In *Snowbird Learning Workshop*, 2004.

Murphy, K. P., Weiss, Y., and Jordan, M. I. Loopy belief propagation for approximate inference: An empirical study. In *UAI*. 1999.

Nivre, J., Hall, J., and Nilsson, J. Memory-based dependency parsing. In *CoNLL*, 2004.

Nivre, J., Hall, J., Kubler, S., McDonald, R., Nilsson, J., Riedel, S., and Yuret, D. The CoNLL 2007 shared task on dependency parsing. In *CoNLL*, 2007.

Riedel, S. Improving the accuracy and efficiency of MAP inference for Markov logic. In *UAI*, 2008.

Riedel, S. and Clarke, J. Incremental integer linear programming for non-projective dependency parsing. In *EMNLP*, 2006.

Riedel, S. and Smith, D. A. Relaxed marginal inference and its application to dependency parsing. In *HLT-NAACL*, 2010.

Smith, D. A. and Eisner, J. Dependency parsing by belief propagation. In *EMNLP*, 2008.

Sontag, D. and Jaakkola, T. New outer bounds on the marginal polytope. In *NIPS*, 2007.

Tromble, R. W. and Eisner, J. A fast finite-state relaxation method for enforcing global constraints on sequence decoding. In *HLT-NAACL*, 2006.

Wainwright, M. and Jordan, M. *Graphical Models, Exponential Families, and Variational Inference.* Now Publishers, 2008.